# Parameter Decoupling Strategy for Semi-supervised 3D Left Atrium Segmentation

Xuanting Hao[*][b], Shengbo Gao[*][a,c,d], Lijie Sheng[b], Jicong Zhang[a,c,d]

[a]School of Biological Science and Medical Engineering, Beihang University, Beijing, China
[b]School of Computer Science and Technology, Xidian University, China
[c]Hefei Innovation Research Institute, Beihang University, Hefei, Anhui, China
[d]Beijing Advanced Innovation Centre for Biomedical Engineering, Beihang University; Beijing Advanced Innovation Centre for Big Data-Based Precision Medicine, Beihang University, Beijing, China

**ABSTRACT**

Consistency training has proven to be an advanced semi-supervised framework and achieved promising results in medical image segmentation tasks through enforcing an invariance of the predictions over different views of the inputs. However, with the iterative updating of model parameters, the models would tend to reach a coupled state and eventually lose the ability to exploit unlabeled data. To address the issue, we present a novel semi-supervised segmentation model based on parameter decoupling strategy to encourage consistent predictions from diverse views. Specifically, we first adopt a two-branch network to simultaneously produce predictions for each image. During the training process, we decouple the two prediction branch parameters by quadratic cosine distance to construct different views in latent space. Based on this, the feature extractor is constrained to encourage the consistency of probability maps generated by classifiers under diversified features. In the overall training process, the parameters of feature extractor and classifiers are updated alternately by consistency regularization operation and decoupling operation to gradually improve the generalization performance of the model. Our method has achieved a competitive result over the state-of-the-art semi-supervised methods on the Atrial Segmentation Challenge dataset, demonstrating the effectiveness of our framework. Code is available at https://github.com/BX0903/PDC.

**Keywords:** Semi-supervised learning, Parameter Decoupling, Consistency Regularization, Left Atrium Segmentation.

## 1. INTRODUCTION

Efficient and accurate segmentation of left atrium (LA) from 3D magnetic resonance (MR) images is of great significance for the treatment of atrial fibrillation. In recent years, deep learning has been used for automatic segmentation on LA and made great progress [1]. In the supervised methods, a large amount of high-quality labeled data plays an important role. However, to delineate numerous reliable pixel-level labels from 3D medical images by experienced exports is expensive and impractical. Therefore, we focus on studying the method of LA segmentation by leveraging rare labeled data with sufficient unlabeled data.

To reduce the labeling demand, various semi-supervised methods have been proposed for medical image segmentation [2,3,4,5,6]. Among these methods, consistency regularization approach was regarded as an effective method, which generally requires models to produce consistent predictions under random perturbation of original images. For example, Yu *et al.* [4] presented an uncertainty aware strategy to mean teacher (MT) framework [7], where the student model learns the uncertainty-guided target from the teacher model. Due to the structure information is ignored, the LG-ER-MT [8] was proposed to force the local and global structural constraint on the different predictions. By exploiting the same basic MT framework, Wang and Zhao *et al.* [9] further explored the feature uncertainty. A multi-task framework was used to capture shape-aware information in [10], which utilized the consistency between pixel-wise classification map and level set representation of the segmentation. While promising progress has been achieved, under the consistency framework, the model parameters for comparative learning would tend to be coupled to output the same result. This would reduce the model's ability to utilize unlabeled data, resulting in a fully supervised model.

---

[*] Both authors contributed equally to this work.
Lijie Sheng: E-mail: ljsheng@xidian.edu.cn
Jicong Zhang: E-mail: jicongzhang@buaa.edu.cn

In this paper, we propose a novel parameter decoupling semi-supervised strategy to solve the aforementioned coupled problem. Motivated by the spirit of [11,12] to construct diverse views, the core idea of our strategy is to allow the network to provide reasonable predictions from different views, which are formed by decoupled parameters. Specifically, we adopt a simple framework that utilizes two output branches to predict with a shared feature extractor. In order to use unlabeled data for regularization, we impose the predictions consistency target on the extractor, aiming to have the same outputs from different perspectives. To further expand the difference in perspectives, we attempt to decouple the parameters of the two branches by the quadratic cosine distance, so that the parameters can be linearly independent to capture different features. Finally, under the alternate optimization of the two operations, the features can be generalized in two perspectives. Our approach was exhaustively evaluated on the dataset of MICCAI 2018 Atrial Segmentation Challenge [13] and achieved a competitive result over the state-of-the-art semi-supervised segmentation approaches.

## 2. METHOD

In this section, we introduce the proposed semi-supervised medical image segmentation framework based on parameter-decoupling consistency. The overall framework is illustrated in Figure 1, which consists of a feature extractor of V-Net architecture and two lightweight classification heads. In the following two subsections, we first introduce the semi-supervised training framework, then introduce the parameter decoupling strategy in detail.

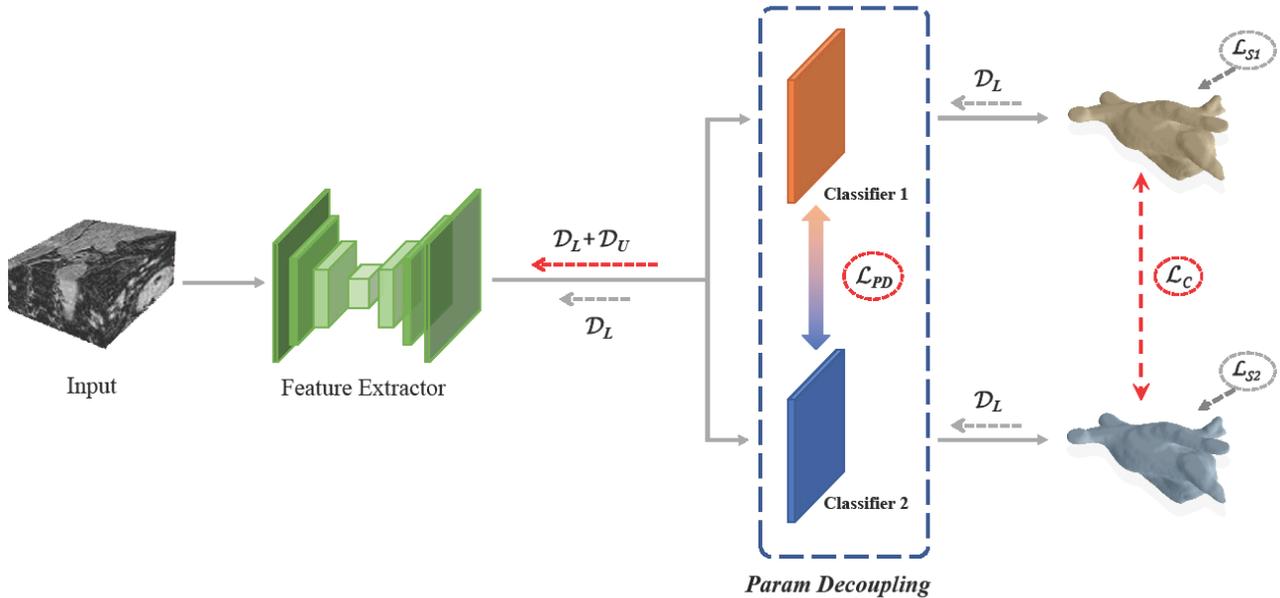

Figure 1. Overview of proposed parameter decoupling framework for semi-supervised segmentation. The network consists of a feature extractor and a dual classifier. The whole framework is optimized by minimizing the supervised loss $\mathcal{L}_S$ on the labeled data $\mathcal{D}_L$. And the consistency loss $\mathcal{L}_C$ on both labeled data $\mathcal{D}_L$ and unlabeled data $\mathcal{D}_U$ is exploited to constraint extractor. The $\mathcal{L}_{PD}$ is used to decouple the parameters of dual classifier.

### 2.1 Semi-supervised Segmentation
#### 2.1.1 Semi-supervised task formulation

We study semi-supervised method by the task of 3D left atrium image segmentation, where the training set consists of $N$ 3D CT scan images with reliable segmentation annotations and $M$ 3D CT scan images without any annotation ($N \ll M$).

We denote the labeled and unlabeled set as $\mathcal{D}_L = \{(x_i, y_i)\}_{i=1}^{N}$ and $\mathcal{D}_U = \{x_i\}_{i=N+1}^{N+M}$ respectively, where $x_i \in R^{H \times W \times D}$ are the input volumes and $y_i \in \{0,1\}^{H \times W \times D}$ are the ground-truth annotations for two classes (e.g., background and left atrium).

### 2.1.2 Segmentation framework

To formulate a 3D semi-supervised segmentation framework, we deploy a V-Net [14] with dual classifier as shown in Figure 1. We denote $f(\cdot)$ as the segmentation framework, $\xi_m$ ($m \in \{1,2\}$), and $\theta$ separately represents the parameter of classifier1, classifier2 and feature extractor. The basic of our semi-supervised segmentation framework is minimizing the following combined supervised objective function:

$$\mathcal{L}_S = \frac{1}{2} \sum_{m=1}^{2} \sum_{i=1}^{N} \mathcal{L}_{Dice}(f(x_i; \theta, \xi_m), y_i) + \mathcal{L}_{CE}(f(x_i; \theta, \xi_m), y_i), \quad (1)$$

where $\mathcal{L}_{Dice}$ and $\mathcal{L}_{CE}$ represent Dice loss and Cross-entropy loss respectively, which measures the quality of classifiers' prediction on labeled images.

Base on the *smoothness assumption*, enforcing multiple models or different branches of a model that both produce consistent predictions for the same image is an effective way of gaining information from unlabeled data. Therefore, we utilize the dual classifier to achieve multi-view consistency construction in latent space by minimizing the loss function on feature extractor:

$$\mathcal{L}_C = \sum_{i=1}^{N+M} \mathcal{L}_{MSE}\big(f(x_i; \theta, \xi_1), f(x_i; \theta, \xi_2)\big), \quad (2)$$

where $\mathcal{L}_{MSE}$ represents the Mean-Square-Error loss function for measuring the consistency between the predictions of dual classifier.

The key challenge of applying consistency loss to regular models is how to avoid the degradation caused by classifier coupling and extract the diversified information from feature space. In order to solve this problem, we propose the parameter decoupling strategy and detailedly describe it as below.

## 2.2 Parameter Decoupling Strategy

### 2.2.1 Parameter decoupling strategy

We notice that under the consistency framework, the dual classifier would tend to be linearly correlated to reach a coupled state. When dual classifiers are coupled, the semi-supervised model will degenerate into a supervised model and lose the capability to exploit unlabeled data for regularization.

To solve this problem, we attempt to decouple the classifiers. To be specific, cosine distance measures the cosine value of the angle between two vectors, which indicates the linear relationship between vectors. As we anticipate that the classifier parameters should be linearly independent in our strategy, the quadratic cosine distance between two classifier parameters is chosen to be the decoupling target:

$$\mathcal{L}_{PD} = \frac{1}{K} \sum_{\substack{\vec{p_1} \in \xi_1 \\ \vec{p_2} \in \xi_2}} \left( \frac{\vec{p_1} \cdot \vec{p_2}}{|\vec{p_1}||\vec{p_2}|} \right)^2, \quad (3)$$

where $\vec{p_1}$ and $\vec{p_2}$ separately represent the flattened parameters of each layer in the dual classifier, and $K$ is the number of parameters of a classifier. The range of the cosine distance is $[-1,1]$, which means that the correlation of vectors changes from negative correlation to positive correlation, and 0 represents the linear independent relationship between vectors. In this paper, we take the quadratic cosine distance as the optimization goal, which makes the classifier parameters orthogonal, increases the view differences of information, and makes the consistency loss function acting on the feature extractor plays a greater role.

### 2.2.2 Overall training process

In the overall training process, the parameters of feature extractor and classifiers are updated alternately by consistency operation and decoupling operation.

Specifically, the supervised loss $\mathcal{L}_S$ guides the two parts of the model to make reasonable predictions on the labeled data. At the same time, under the influence of decoupling loss $\mathcal{L}_{PD}$, the classifiers become orthogonal. As a consequence, the dual classifier can be viewed as two distinct classification hyperplanes, which implement classification based on different components of the relatively redundant features provided by the extractor. Then, on the basis of decoupled classifiers, the feature extractor is forced to generate features that make the output of two classifiers the same through consistency loss $\mathcal{L}_C$. Although the training process does not explicitly optimize a typical minimax problem, rather optimize a pair of tasks with conflicting results. Therefore, we call it an implicit adversarial pipeline. In the end, the region that the classification boundary passes through is sparser, which makes the generalization ability of the feature extractor improved.

## 3. EXPERIMENTS AND RESULTS

### 3.1 Dataset and Pre-processing

Our method is evaluated on the Atrial Segmentation Challenge dataset. It consists of 100 3D gadolinium-enhanced MR imaging scans (GE-MRIs) and ground truth segmentation masks, with an isotropic resolution of $0.625 \times 0.625 \times 0.625 mm^3$. In this paper, the experiment following [4], 80 scans are used for training and 20 scans are used for testing. All the scans are cropped centering at the heart region and standardized.

### 3.2 Implementation Details

The framework is implemented based on PyTorch, using an NVIDIA Tesla V100 GPU. We randomly crop $112 \times 112 \times 80$ sub-volumes and use the same data augmentation techniques as the [15]. The batch size is 4, half of which are annotated images and the rest is unannotated images. The whole framework is trained by the SGD optimizer for 6000 iterations. The initial learning rate is 0.01, which is divided by 10 every 2500 iterations. We set $\lambda_C$, $\lambda_{PD}$ as time-dependent Gaussian warming up function $\lambda(t) = 0.1 * e^{(-5(1-t/t_{max})^2)}$, which are multiply $\mathcal{L}_C$ and $\mathcal{L}_{PD}$ to balance the three loss terms in training process respectively.

### 3.3 Evaluation and Comparison

Table 1. Quantitative comparison between our method and various semi-supervised methods on the Atrial Segmentation Challenge dataset. The last three rows are our ablation experiment results.

| Method | Scans used | | Metrics | | | | | |
|---|---|---|---|---|---|---|---|---|
| | Labeled | Unlabeled | Dice(%) | Jaccard(%) | ASD(voxel) | 95HD(voxel) | CD | QCD |
| V-Net | 80 | 0 | 91.14 | 83.82 | 1.52 | 5.75 | - | - |
| V-Net | 16 | 0 | 86.03 | 76.06 | 3.51 | 14.26 | - | - |
| MT [7] | 16 | 64 | 88.23 | 79.29 | 2.73 | 10.64 | 0.9998 | 0.9996 |
| UA-MT [4] | 16 | 64 | 88.88 | 80.21 | 2.26 | 7.32 | 0.9998 | 0.9997 |
| SASSNet [16] | 16 | 64 | 89.27 | 80.82 | 3.13 | 8.83 | -0.9934 | 0.9870 |
| DTC [10] | 16 | 64 | 89.42 | 80.98 | 2.10 | 7.32 | -0.8019 | 0.6823 |
| DUA-MT [9] | 16 | 64 | 89.65 | 81.35 | 2.03 | 7.04 | - | - |
| V-Net-GC | 16 | 64 | 87.52 | 78.29 | 4.10 | 14.24 | 0.9150 | 0.8374 |
| V-Net-EC | 16 | 64 | 87.85 | 78.74 | 2.60 | 8.19 | 0.9045 | 0.8182 |
| **PDC-Net** | 16 | 64 | **89.76** | **81.57** | 2.95 | 10.31 | **0.4366** | **0.3097** |

Our method is evaluated on four metrics, *i.e.*, Dice, Jaccard, Average Surface Distance (ASD) and 95% Hausdorff Distance (95HD). Besides, we apply cosine distance (CD) and quadratic cosine distance (QCD) to measure the coupling degree in the open-source methods. It is worth emphasizing that all the methods are evaluated by the same backbone.

We firstly present the full-supervised segmentation results which are trained on 80 and 16 images as the upper-bound and the lower-bound of this task. In order to make a fair comparison, we follow the setting of previous works, which utilize 20% of the training set as labeled data and the rest 80% images as unlabeled data for the semi-supervised task.

It can be observed from Table 1 that previous semi-supervised methods have achieved excellent results on this task, but CD and QCD indicate that there may be coupling problems in these methods, which limits their ability to use unlabeled data. The self-ensemble-based methods MT [7] and UA-MT [4] perform the highest coupling degrees, which are evaluated between the student model and the teacher model. The SASSNet [16] and DTC [10] utilize data by building a multi-task network and also meet the coupling problem, the CD and QCD are evaluated between the different branches in the framework. By comparing with the above methods, our PDC-Net (Parameter Decoupling Consistency Net) achieved better segmentation results under a simple framework with parameter decoupling in both Dice (89.76%) and Jaccard (81.57%) under the lowest CD and QCD. However, in order to show the improvement that is completely brought about by PDC, the PDC-Net is the direct combination of V-Net and PDC. In addition to MT, previous methods in Table 1 merged shape regression or uncertainty awareness to impose stronger boundary constraints. Therefore, PDC-Net did not achieve the best performance on shape-aware metrics (ASD and 95HD). Some qualitative results are provided in Figure 2.

To analyze the effectiveness of our parameter decoupling strategy and quantify the performance of each component in our framework, we present the ablation experiment on the following models: 1) V-Net-GC: V-Net with global consistency which optimizes supervised loss and consistency loss for the entire network at the same time. 2) V-Net-EC: V-Net with extractor consistency which firstly optimizes supervised loss for the entire network, then optimize consistency loss for extractor alone in each iteration. and 3) PDC-Net: The whole framework. It can be observed that the extractor consistency outperforms the global consistency, and parameter decoupling strategy can greatly improve the performance (2.24% in Dice, 3.28% in Jaccard).

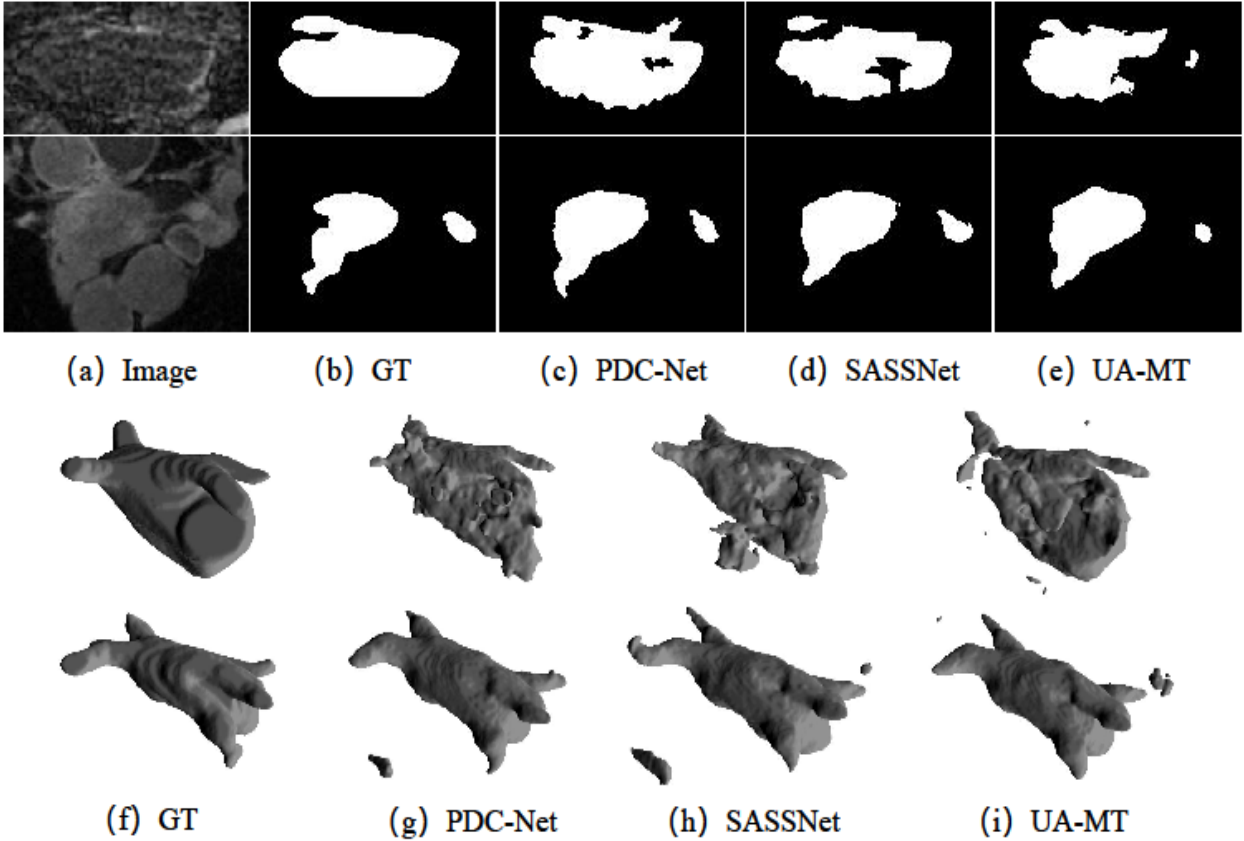

Figure 2. 2D visualization (a-e) and 3D visualization (f-i) of the segmentations from different methods.

## 3.4 Analysis of the Amount of Annotated Data

Table 2. Comparison between PDC-Net and V-Net-GC under mutiple data volume settings.

| Method | Scans used | | Metrics | | | |
|---|---|---|---|---|---|---|
| | Labeled | Unlabeled | Dice(%) | Jaccard(%) | ASD(voxel) | 95HD(voxel) |
| V-Net-GC | 8 | 72 | 83.88 | 72.87 | 4.17 | 15.71 |
| **PDC-Net** | 8 | 72 | **86.55** | **76.57** | 3.92 | 13.60 |
| V-Net-GC | 16 | 64 | 87.52 | 78.29 | 4.10 | 14.24 |
| **PDC-Net** | 16 | 64 | **89.76** | **81.57** | 2.95 | 10.31 |
| V-Net-GC | 24 | 56 | 90.06 | 82.02 | 2.53 | 7.36 |
| **PDC-Net** | 24 | 56 | **90.33** | **82.44** | 2.42 | 8.91 |

To highlight the role of parameter decoupling strategy, we compare PDC-Net with V-Net-GC under extra data volume settings, where 10% and 30% of the training set are employed as the labeled set. In Table 2, it is obvious that the performance gap between PDC-Net and V-Net-GC gradually increasing with the decrease of labeled data ratio (Dice: $0.27\% \rightarrow 2.24\% \rightarrow 2.67\%$), which supports the effectiveness of parameter decoupling strategy in improving the performance of the consistency-based method.

## 4. CONCLUSION

In this paper, we proposed a simple yet novel parameter decoupling semi-supervised segmentation framework for 3D left atrium MR images. In contrast to previous consistency-based methods, our method exploits parameter decoupling of classifiers to construct diverse views of latent feature space for enforcing feature extractor to improve the generalization performance. The comparison with other semi-supervised methods demonstrates the phenomenon we observed widely exists in consistency-based methods and can be mitigated effectively. The future works include investigating the effect of different parameter decoupling manners and extending our framework to other semi-supervised medical image segmentation problems.